# Shortcut Learning in Legal Judgment Prediction: Empirical Evidence from the UK Employment Tribunal

**Joe Watson[1,2*], Joana Ribeiro de Faria[1], Marcus Tomalin[3], Måns Magnusson[4], Huiyuan Xie[5], Hao Tian Yeung[6], Christine Carter[1], Jonathan Rutherford[1], Felix Steffek[1]**

1. Faculty of Law, University of Cambridge, Cambridge, United Kingdom
2. The Psychometrics Centre, Cambridge Judge Business School, University of Cambridge, Cambridge, United Kingdom
3. Faculty of English, University of Cambridge, Cambridge, United Kingdom
4. Department of Statistics, Uppsala University, Uppsala, Sweden
5. Department of Computer Science and Technology, Tsinghua University, Beijing, China
6. Department of Engineering, University of Cambridge, Cambridge, United Kingdom

* Corresponding author: Joe Watson, jmw239@cam.ac.uk

## Abstract

Current Legal Judgment Prediction (LJP) is constrained by its reliance on post-hoc judicial materials, increasing the likelihood that models perform retrospective classification rather than true forecasting. This paper empirically investigates shortcut learning in this context by studying claim-level outcome prediction in UK Employment Tribunal (UKET) decisions. Using a corpus of 33,158 individual claims, we predict outcomes from claim texts and LLM-extracted case summaries, evaluating models ranging from interpretable TF-IDF-based classifiers to black-box LLMs. While headline predictive performance figures appear strong, we demonstrate that such performance in LJP systems trained on post-hoc judicial text can be driven by the retrospective nature of the source material. Stratifying the test data by human judgments of leakage reveals that performance increases where outcome-revealing cues are embedded in the narrative. Moreover, a model trained on just the 4% of features identified as leakage achieves high performance, outperforming human experts. These findings substantiate concerns that LJP performance may be exaggerated by linguistic artefacts. Yet this vulnerability is not fatal to the research agenda. Instead, post-hoc judgments might be treated as potentially contaminated texts, requiring active auditing. Retraining models after masking leakage features results in only a negligible reduction in Macro-F1. Hence, while models will opportunistically exploit shortcuts when available, they remain capable of extracting useful predictive signals when these artefacts are removed.

# Main Text

## 1 Introduction

The practical promise of Legal Judgment Prediction (LJP) rests on the possibility of forecasting court outcomes from information available before a decision is made, estimating a dispute's likely resolution from ex ante case materials. In its strongest form, therefore, LJP is a temporally constrained task: a model should not rely on information that would only become available after the relevant legal decision has been reached. This requirement matters because LJP is often motivated by its potential to help litigants, lawyers, or policymakers assess likely outcomes before proceedings. For that promise to hold, predictions must be based on information available at that point, rather than on facts, findings, or reasoning later used to justify the result.

This ideal is difficult to satisfy in practice because most LJP work relies on post-hoc judicial texts: decisions published after the court outcome has been reached. This reliance largely arises from data constraints. Pre-decisional materials, such as claim forms, response forms, comprehensive case files, and other pre-trial documents, are rarely available to researchers (Medvedeva et al. 2023), including in the context of the UK Employment Tribunal (UKET). As a result, many LJP studies risk performing outcome identification or outcome classification rather than true forecasting (Medvedeva et al. 2023). The problem is that judicial decisions are not neutral repositories of ex ante facts (Liu et al. 2025). They may contain facts, findings, and reasons selected and presented in light of the outcome itself (Leubsdorf 2001). Models trained on such texts may therefore appear to predict legal outcomes while, in fact, relying on information that was not available at the time when prediction would have been useful.

Early LJP research focused primarily on whether computational models could predict outcomes from accessible legal texts, often reporting strong performance. Studies using various forms of machine learning (ML) achieved high performance across diverse systems, including the European Court of Human Rights (ECHR), the Supreme People's Court of China, and the Supreme Court of India. For instance, neural network models have set strong baselines for ECHR cases (Chalkidis et al. 2019), and large-scale datasets such as China's CAIL2018 have enabled models to assign legal labels to factual descriptions with high precision (Xiao et al. 2018). Similar benchmarks have been reached in India using the eLegPredict model and the Indian Legal Documents Corpus (Sharma et al. 2022; Malik et al. 2021). However, strong benchmark performance is not itself evidence of genuine legal forecasting. As argued by Medvedeva and McBride (2023), relatively few LJP studies can be strictly characterised as predicting future judgments.

The underlying validity problem can be understood as a form of shortcut learning. Shortcut learning occurs when a model achieves high performance by exploiting statistical correlations that are not causally or conceptually related to the task it is supposed to solve (Geirhos et al. 2020; Bihani and Rayz 2025). In LJP, this risk is especially acute because the available modelling data are often confined to "highly selective [case] summaries" tailored to support a judicial decision (Tippett et al. 2022, p 1162). Training on such data might mean that models merely learn retrospective rationalisations rather than true predictive features (Pasquale and Cashwell 2018). Outcome-revealing cues may appear directly, as where a judgment states that a claim “succeeds” or “fails”, or indirectly, through retrospective judicial language such as findings that evidence was “not accepted” or that a party “failed to establish” an allegation. This creates a “Clever Hans” effect, in which a model appears intelligent but actually relies on unintended cues embedded in the data (Lapuschkin et al. 2019).

This shortcut-learning concern connects directly to a broader criticism of LJP’s lack of chronological integrity. In a valid predictive setting, models must only use inputs available at the time of prediction. Researchers have identified temporal leakage across a large number of studies, showing that judicial texts are systematically problematic as predictive inputs because they are written after the fact (Medvedeva et al. 2023; Sargeant and Magnusson 2024). These critiques suggest that headline LJP performance figures may overstate what models could achieve in real pre-decisional settings.

A logical response is to substitute the source data used in LJP tasks, replacing post-hoc judgments with pre-decisional material. Such an approach was adopted by Medvedeva and colleagues (2021) as an intervention, to explore the impact of data leakage on predictive performance. Their findings show that performance drops when models are trained on genuine ECHR pre-judgment documents rather than post-hoc judgment summaries. This is a valuable intervention because it demonstrates empirically that the choice of data source materially affects LJP performance. However, it is not an intervention that can be applied more broadly. Pre-decisional materials are rarely available at scale, particularly in the UK legal system. Moreover, where such materials are available, they may not include all information that ultimately informs the judgment. Performance under the pre-decisional condition may therefore be reduced both because leakage has been removed and because legally relevant input information is incomplete.

This paper addresses that constraint directly. Rather than treating leakage as a binary property of an entire dataset, we investigate how leakage appears within post-hoc judicial texts, how much it affects model performance, and whether it can be reduced through data modification. The central contribution is therefore not merely another LJP benchmark. It is a leakage-centred evaluation of LJP under realistic data constraints. We operationalise shortcut learning by identifying outcome-revealing linguistic cues, testing the extent to which models rely on them, and evaluating whether predictive performance changes when those cues are removed. This approach follows NLP research showing that models can be “right for the wrong reasons” by exploiting superficial dataset artefacts rather than learning the intended task (Gururangan et al. 2018; McCoy et al. 2019; Poliak et al. 2018).

The UK Employment Tribunal provides a strong setting for this inquiry. UKET produces a high volume of publicly available, English-language judgments, many of which are available in the Cambridge Law Corpus (CLC, Östling et al. 2023). These judgments are also highly granular. A single decision may resolve multiple distinct claims, such as unfair dismissal, discrimination, unlawful deduction from wages, or breach of contract. This structure allows leakage to be examined at the level at which outcomes are actually assigned, rather than through a single aggregate case label: claim-level prediction is used here as a methodological device for diagnosing shortcut learning in post-hoc judgments.

This claim-level design also produces an ancillary benefit by addressing a gap in UK LJP: to our knowledge, no prior academic work has modelled the outcomes of individual claims within UK judgments. Existing research on UK case outcome prediction remains nascent and has so far been restricted to aggregate, case-level labels (e.g. Xie et al. 2024). Still, we acknowledge that fine-grained outcome prediction has taken place outside the UK. Related work has explored article-level outcome prediction in ECHR cases (Aletras et al. 2016; Chalkidis et al. 2019), multi-charge criminal cases framed as multi-label prediction (Li et al. 2025), and appellate legal prediction in civil cases using datasets that include claim-level decision annotations (Huang et al. 2026). The novelty of the present study lies in applying this level of granularity to UK employment tribunal judgments, where it is used to diagnose shortcut learning in post-hoc judicial texts.

The paper's main contribution is to show how shortcut learning can arise in LJP models trained on post-hoc judicial texts, how strongly such models may rely on outcome-revealing linguistic cues, and whether useful predictive signal remains once those cues are removed. We examine this problem in UKET judgments by identifying retrospective cues, measuring their effect on model behaviour, and testing whether performance changes when those cues are masked. Our findings show that models exploit shortcut cues when available yet continue to perform reasonably well when the identified leakage features are removed during training. This suggests that post-hoc judicial texts need not be discarded wholesale, but should be subjected to active auditing before they are used for prediction tasks. A claim-level variant of the CLC UKET dataset provides the empirical basis for this analysis, and is released alongside associated resources, including claim-level outcome labels, human judgments of leakage, and human outcome baselines.

**Research Question**
The paper is guided by the following overarching research question:

RQ: To what extent is the predictive success of LJP models driven by shortcut learning cues embedded in post-hoc judicial texts?

We examine this question in two stages. First, we test whether LJP models perform better when outcome-revealing cues are present and whether such cues exert substantial influence on model behaviour. Second, we test whether predictive performance persists after identified shortcut learning cues are removed.

# 2 Methods

## 2.1 Dataset and Data Preparation

The data (Table 1) are derived from written decisions of the UK Employment Tribunal (UKET). These public documents record the Tribunal's reasoning and final decision for cases that frequently involve multiple, legally distinct claims. As these are post-hoc judicial texts written to justify a determined outcome, they inherently risk containing predictive "leakage" that may artificially inflate model performance.

**Table 1. Data Descriptive Statistics**

| Category | Statistic | Value |
|---|---|---|
| Totals | Total Judgments (Cases) | 13,253 |
| | Total Claims | 33,158 |
| | Avg. Claims per Judgment | 2.50 |
| Train/Test Split | Training Set (LLM-annotated) | 32,030 (96.6%) |
| | Test Set (Human-annotated) | 1,128 (3.4%) |
| Annotator Distribution | Expert-Annotated Ground-Truth Labels | 1,128 |
| | LLM-Annotated Training Labels | 32,030 |
| Class Distribution (Overall) | Win | 13,104 (39.5%) |
| | Loss | 16,174 (48.8%) |
| | Other / Procedural | 3,880 (11.7%) |
| Leakage (Test Set Only) | Claims marked as leakage by human expert | 289 (25.6%) |
| Text Length (Tokens) | Claim Summaries (Mean / Median / Range) | 5.6 / 4.0 / 1–30 |
| | Case Facts (Mean / Median / Range) | 80.7 / 80.0 / 2–462 |

*Note*. A claim marked as leakage was labelled as such by at least one of two annotators (see Leakage Annotations).

To prepare the data for modelling, we processed each judgment into three primary components:

- **Case Facts:** We used pre-existing summaries (developed for a previous analysis of UKET data; Xie et al. 2024) to present the facts underlying each case. The original prompting for these summaries was designed to exclude outcome statements, though the post-hoc nature of the source text means that subtle cues may remain.
- **Claim Descriptions:** For each individual claim within a judgment, we used DeepSeek (v3.1)[1] to extract a short textual description (e.g. "discrimination due to age"). These descriptions, paired with case-fact summaries, form the textual input for our models.
- **Outcome Labels**: Each claim was assigned a label of Win, Loss, or Other. Other means an outcome that cannot be classified as Win or Loss, such as an order to submit more evidence.

## 2.2 Annotation and Labelling Strategy

To scale training without the prohibitive cost of expert manual coding, we adopted a weak-supervision framework (Ratner et al. 2017). We used an LLM to extract 32,030 labels from the tribunal judgments to serve as the training set. While these automated labels contain inherent parsing noise, recent evidence suggests that LLM-based annotation can effectively substitute for traditional crowd-working at scale (Gilardi et al. 2023). Following the 'Train on Synthetic, Test on Real' (TSTR) paradigm, model performance was evaluated exclusively against Expert-Annotated Ground-Truth Labels for 1,128 claims: claim outcomes manually annotated by a human expert with access to the full, unrestricted judgment text. For modelling and evaluation, outcome categories were collapsed into three labels: Win, Loss, and Other. The gold annotation guidance used to create these ground-truth labels, which also shows the mapping between collapsed labels and more granular outcomes, is provided in SI 1.

[1] DeepSeek was chosen because it provided a comparatively low-cost model for large-scale annotation, offered sufficiently reliable structured outputs in preliminary testing, and was available as an open-weight model, supporting reproducibility and potential local deployment.

To provide a human comparator under identical informational constraints, we established a Matched-Input Human Benchmark. The Matched-Input Human Benchmark consists of human outcome predictions from trained law PhD students with a specialisation in employment law who were provided with the same restricted informational inputs as the models: case summaries and claim descriptions. The guidance shown to these annotators, including the instructions for outcome prediction and leakage identification, is provided in SI 2. This allows us to compare model performance both against the Expert-Annotated Ground-Truth Labels and against human experts operating under the same informational constraints as the models. We also maintained a judgment-level split, preventing cross-document contamination by ensuring that no claims from the same case appeared in both the training and test sets. Finally, the human expert annotators reviewed the test set to identify outcome-revealing language, facilitating the analysis of shortcut learning by distinguishing between explicit and implicit leakage.

## 2.3 Modelling and Evaluation

We evaluated three model classes across the complexity-interpretability spectrum. First, glass-box linear models (Logistic Regression and Linear SVM) were trained on TF-IDF features to allow for direct coefficient inspection. Second, a transformer-based approach used a frozen DistilBERT sentence-embedding model (all-MiniLM-L6-v2), with the resulting embeddings classified by a linear SVM. Third, DeepSeek was evaluated as a generative classifier using zero-shot and few-shot prompting. All models were evaluated using Macro-F1 to account for class imbalance (Win: 39.5%; Loss: 48.8%; Other: 11.7%).

For the TF-IDF models, all strings were lowercased before vectorisation. We evaluated unigram, bigram, and trigram configurations, restricting the vocabulary to the top 5,000 features by frequency. The trigram (3,3) configuration emerged as the most effective during cross-validated grid search and served as the baseline for all subsequent interpretability analyses. For the DistilBERT model, each claim-level instance was represented by concatenating the claim text with the associated case facts after text cleaning. The frozen encoder generated L2-normalised sentence embeddings, which were then used to train a LinearSVC classifier with balanced class weights.

For the DeepSeek models, each prediction was made for one claim at a time, using the same restricted inputs provided to the supervised models: the LLM-extracted claim description and the case-fact summary. The model was instructed to predict the claim outcome from the claimant's perspective and to return exactly one of three labels: Win, Lose, or Other. In the zero-shot condition, no labelled examples were provided. In the few-shot condition, the prompt included five labelled training examples sampled once from the training pool using a fixed random seed and then reused for all predictions; these examples covered all three outcome classes. Generation was deterministic, with temperature set to 0. Prompts used for zero- and few-shot classification, together with prompts for claim extraction, outcome labelling, and leakage-feature auditing, are available in the project GitHub repository (https://github.com/JoeMarkWatson/uket_pred3).

## 2.4 Diagnostic Analysis of Shortcut Learning

To examine shortcut learning, we conducted a multi-stage diagnostic analysis. Initially, we partitioned the test set into Clean (no apparent leakage) and Leaky (explicit or implicit leakage) subsets based on the human annotations previously described. By calculating the performance differential between these subsets, or "leakage gap", we identified whether model performance was inflated by outcome-revealing cues. To move beyond aggregate metrics, we inspected the internal decision-making of the best-performing linear model by extracting feature coefficients. We applied Kernel Density Estimation (KDE) to compare the weight distributions of leakage versus non-leakage features, identifying whether the decision boundary was disproportionately driven by high-signal shortcut cues.

To isolate specific linguistic artefacts, we conducted a systematic audit of the 5,000 trigram features. We employed DeepSeek as an independent classifier to categorise each trigram as explicit leakage, implicit leakage, or no leakage based on whether the phrase plausibly disclosed the adjudicative resolution (blind to model coefficients and ground-truth labels). This audit identified a Contaminated Feature Set of 200 trigrams (4% of the feature space) comprising procedural giveaways (e.g. "sought to amend") and outcome-implying narrative markers (e.g. "respondents failed to"). The reliability of this feature set was identified through human expert review.

A law professor and a law PhD candidate independently carried out blind validation on a stratified random sample of 30 features (10 per LLM-predicted category). Each trigram feature was assessed without access to LLM classifications, model coefficients, or outcome labels. For intervention purposes, categories were collapsed into a binary distinction: leakage (explicit or implicit) versus no leakage. The reviewers achieved 73% binary agreement, indicating moderate agreement and reflecting the subjectivity of leakage identification. The LLM showed 67% agreement (accuracy) with the PhD candidate and 53% with the professor. Despite this variation, the results suggest the LLM-generated Contaminated Feature Set is sufficiently enriched with outcome-revealing patterns to serve as a proxy for shortcut features – ablation experiments require substantial coverage of high-signal cues rather than perfect precision.

## 2.5 Intervention Experiments

To test for mechanistic reliance on shortcut cues, we conducted three controlled intervention experiments. Each intervention was designed to test a different aspect of leakage dependence: whether leakage features are sufficient for prediction, whether models rely on them at inference time, and whether predictive performance can be recovered when these features are removed before training. Hyperparameters were re-tuned for each retrained condition.

**Leakage-Only Model: Sufficiency**

The Leakage-Only Model tests whether outcome-revealing features alone are sufficient to support legal outcome prediction. We trained a logistic regression classifier exclusively on the 200 trigrams in the Contaminated Feature Set, with all other tokens removed. High predictive performance in this condition would show that LJP can be performed, at least in part, through superficial pattern matching on leakage features alone.

**Test-Set Ablation: Inference-Time Reliance**

The Test-Set Ablation tests whether a model trained on full text relies on leakage features at the point of prediction. We used the original model, trained on the complete feature set, but evaluated it on a modified test set in which the Contaminated Feature Set was removed. This isolates the contribution of leakage at inference time. A substantial performance drop would indicate that the model had learned to depend on outcome-revealing cues when making predictions.

**Training-Set Ablation: Recoverability**
The Training-Set Ablation intervention tests whether useful predictive signal remains recoverable when leakage features are removed before learning. We retrained the model from scratch after removing the Contaminated Feature Set from the training corpus. This prevents the model from learning associations between these specific shortcut features and legal outcomes. If strong performance is maintained in this condition, it suggests that models can still extract legally relevant signal from post-hoc texts even when more obvious leakage cues are unavailable.

Together, these interventions allow us to explore three possibilities: that leakage features are sufficient to support prediction, that models trained on unmodified post-hoc texts rely on such features at inference time, and that valid predictive signal remains recoverable once these shortcuts are suppressed.

# 3 Results

## 3.1 Headline Performance

We first calculated headline performance metrics for all evaluated systems (Fig. 1). Among these, the supervised systems with tuned hyperparameters (TF-IDF plus SVM, and DistilBERT plus SVM) achieved the highest observed scores (up to 0.623), compared with lower scores for the zero-shot and few-shot LLMs (0.47 and 0.54, respectively). All computational models also scored above the random baseline and the Matched-Input Human Benchmark (~0.51). The Matched-Input Human Benchmark was derived from two legal experts who were given exactly the same inputs as the models: LLM-extracted claim descriptions and case-fact summaries. On the surface, these results suggest that various computational models can predict claim outcomes from restricted case summaries and claim descriptions. However, this aggregate view does not reveal whether performance is driven by legally relevant reasoning or by shortcut cues embedded in post-hoc texts.

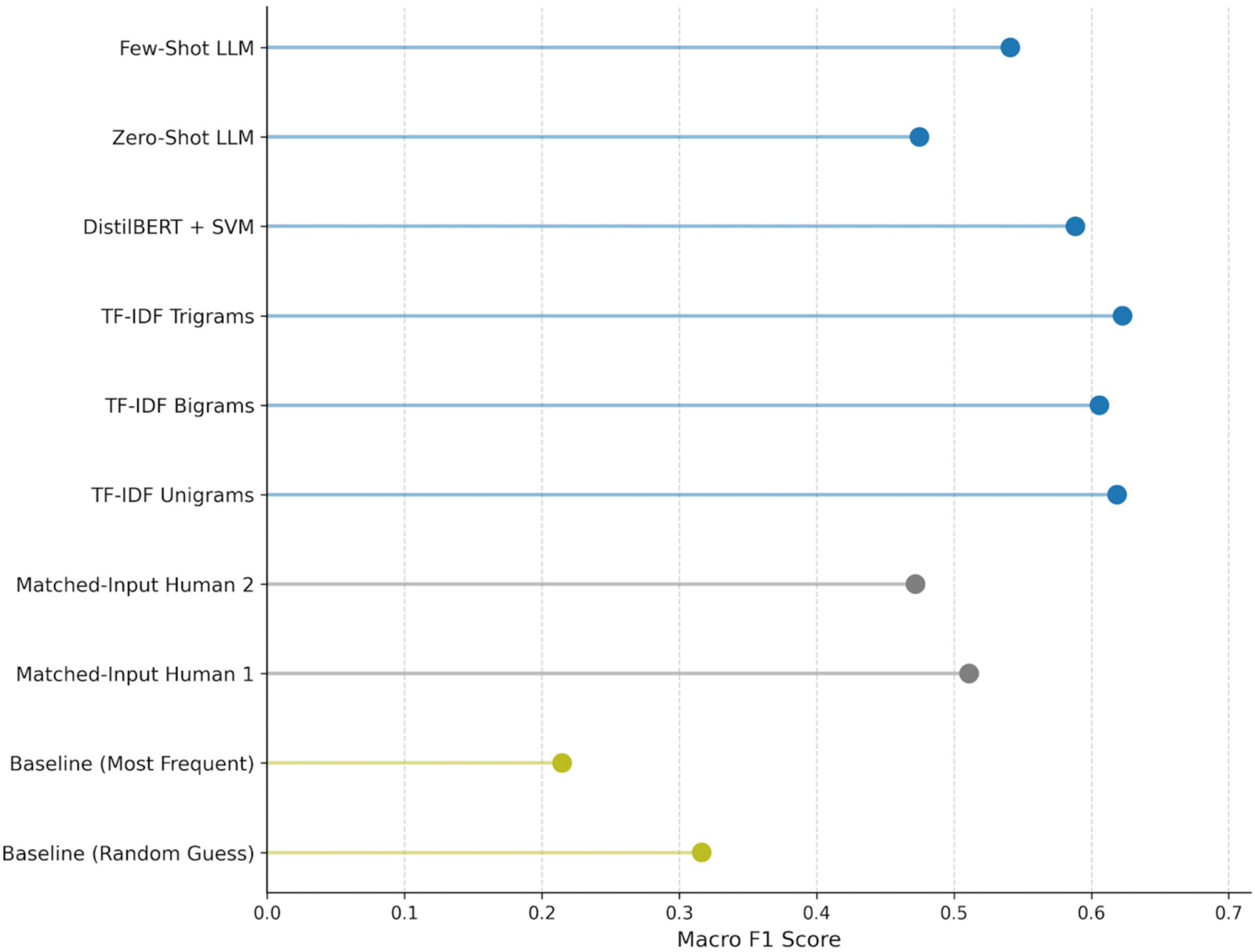

**Fig. 1** Overall Model Performance vs. Baselines and Matched-Input Human Benchmarks. *Note*. This figure presents the predictive performance (Macro-F1) of all evaluated systems against naïve baselines and two Matched-Input Human Benchmarks.

## 3.2 Model Performance on Leakage and Non-Leakage Claims

While headline performance figures suggest that models can predict claim outcomes using case facts, this aggregate view does not reveal whether performance is driven by legally relevant reasoning or by shortcut cues embedded in post-hoc texts. A clear leakage gap emerges for all but one of the standard classifiers. Predictive performance is substantially higher for claims where annotators identified outcome-revealing cues (triangle points, Fig. 2) than for those where no leakage was found (circle points, Fig. 2). This discrepancy suggests that these models actively exploit leakage artefacts to achieve their inflated headline metrics.

The TF-IDF Unigram model is the sole exception to this trend; it does not exhibit a positive leakage gap (-0.01 difference). This likely reflects its structural limitation: unigram models cannot capture multi-word constructions that typically encode leakage (e.g. “judgment for claimant”, “was unfairly dismissed”). Such phrases become predictive only when tokens are considered jointly, a capability available to Bigram, Trigram, and Transformer-based models.

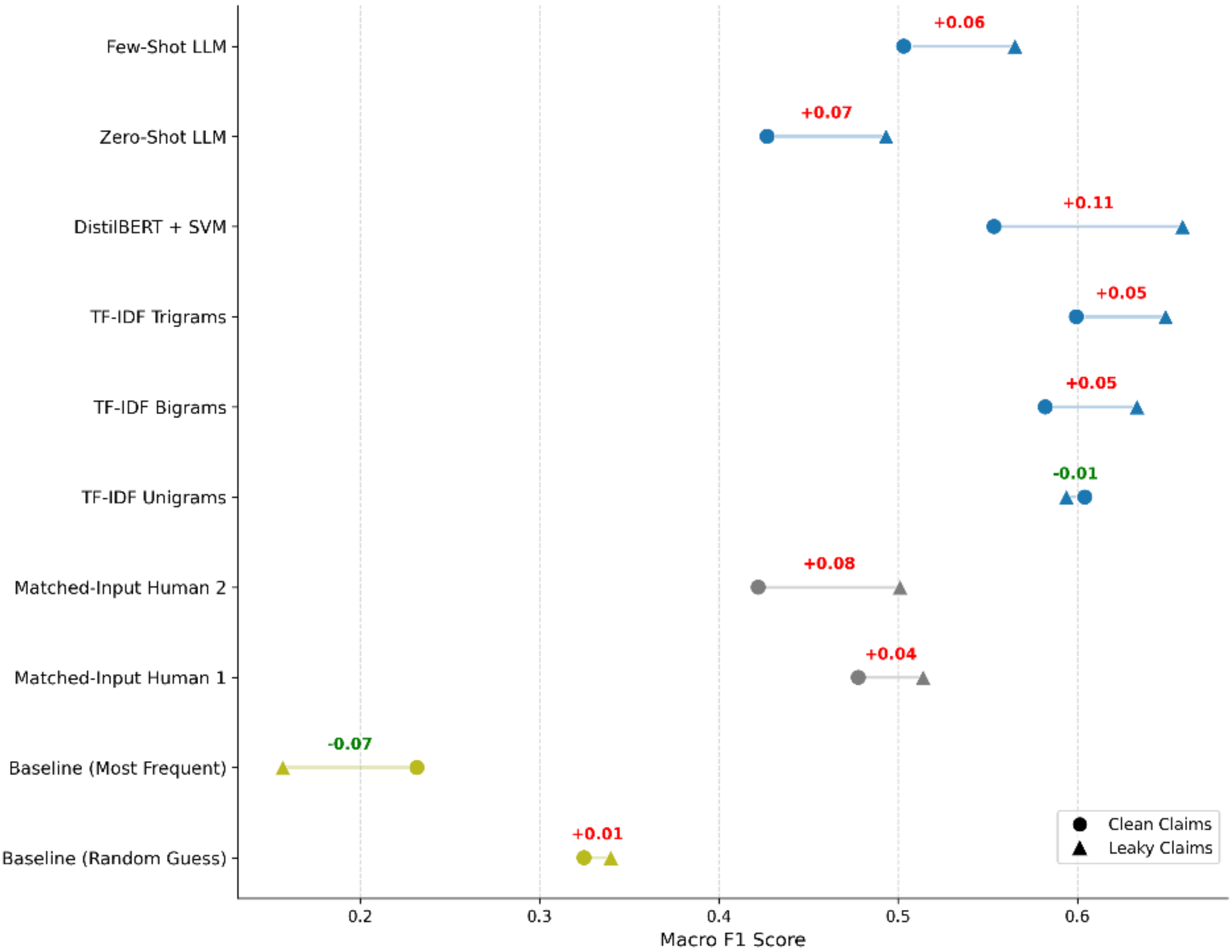

**Fig. 2** Leakage Gap: Main Models (Clean vs. Leaky Claims)

## 3.3 Mechanistic Evidence

The leakage gap suggests that models perform better when outcome-revealing language is present. The feature-weight analysis shows how this happens inside the best-performing interpretable model. In the TF-IDF Trigram model, leakage features are not merely present in the vocabulary. They receive disproportionately large coefficients, meaning that they exert greater influence on the classifier's predictions than most non-leakage features. This provides mechanistic evidence that the model's decision boundary is shaped by shortcut cues.

Figure 3 explores this pattern by comparing the coefficient distributions for features identified by the automated audit as containing leakage with those classified as non-leakage. The non-leakage distributions are generally concentrated around zero, indicating that most ordinary legal and factual language has limited influence on the final classification. By contrast, the leakage distributions are somewhat wider, with heavier tails, showing that outcome-revealing trigrams receive more high-magnitude weights. Examples include phrases such as "unauthorised deduction from" for Win predictions and "claimant failed to" for Loss predictions. The model, therefore, does not merely benefit from leakage in aggregate; it assigns some of its strongest predictive weights to a small set of giveaway phrases.

The pattern is more nuanced for the Other category. In the coefficient density plot, leakage and non-leakage features appear more similar (Fig. 3C) than they do for Win and Loss predictions (Fig. 3A, B), suggesting that leakage does not affect the distribution as clearly. However, the extreme-coefficient analysis shows that leakage features are still more likely than non-leakage features to receive very large weights (Fig. 3F). This indicates that, even where the average distributional difference is less visually pronounced, a subset of leakage features remains highly influential for predicting Other outcomes. Procedural language could potentially be more heterogeneous than language associated with claim wins or losses. That is, some features classified as leakage may be weak predictors, despite a small number of others providing strong procedural giveaways. Overall, the coefficient analysis supports the same conclusion as the leakage-gap results: where outcome-revealing artefacts are available, the model gives them substantial predictive weight.

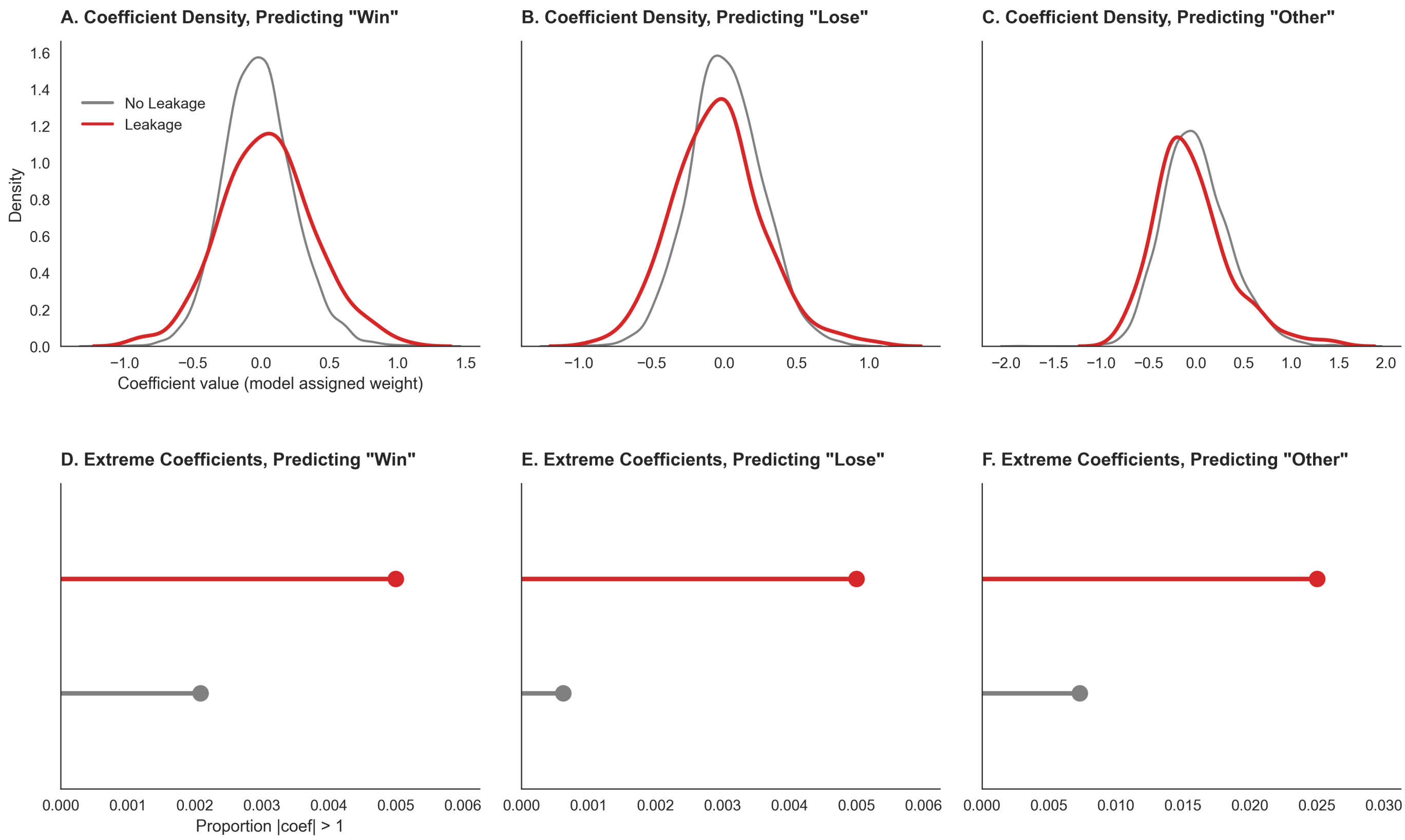

**Fig. 3** Distribution of Feature Weights (Coefficients) in the TF-IDF Trigram Model, Stratified by Leakage Status. *Note*. Panels A–C display kernel density estimates of coefficient magnitudes. Panels D–F report the proportion of coefficients exceeding 1. 'Leakage' (Red) comprises features identified by the LLM audit as containing explicit or implicit outcome cues; 'No Leakage' (Grey) comprises all other features.

## 3.4 Intervention Outcomes

The results of the targeted interventions confirm the role of these shortcuts while highlighting the recoverability of the legal signal (Fig. 4).

**The Leakage-Only Model: Sufficiency**

The Leakage-Only model, trained exclusively on the 200 identified leakage trigrams, achieved a Macro-F1 of 56.07. This substantially outperforms the most-frequent-class baseline and approaches the performance of the full-text models. This result demonstrates that the leakage features alone possess high informational density; a model requires only shortcut cues to achieve strong predictive performance.

**Test-Set Ablation: Inference-Time Reliance**

When the standard model (trained on the full text) was evaluated on a test set where the 200 leakage trigrams were masked, Macro-F1 dropped from 62.27 to 60.85. This performance degradation indicates mechanistic reliance: standard models trained on post-hoc texts learn to treat outcome-revealing cues as important predictors.

**Training-Set Ablation: Recoverability**
Crucially, however, the model proved robust when the shortcuts were removed from the learning process. Retraining the model architecture from scratch while masking leakage cues resulted in a Macro-F1 of 61.62 – only marginally below the 62.27 achieved by the original, contaminated trigram approach. This suggests that while models will opportunistically exploit shortcuts when available, they remain capable of extracting genuine legal signals when these artefacts are removed.

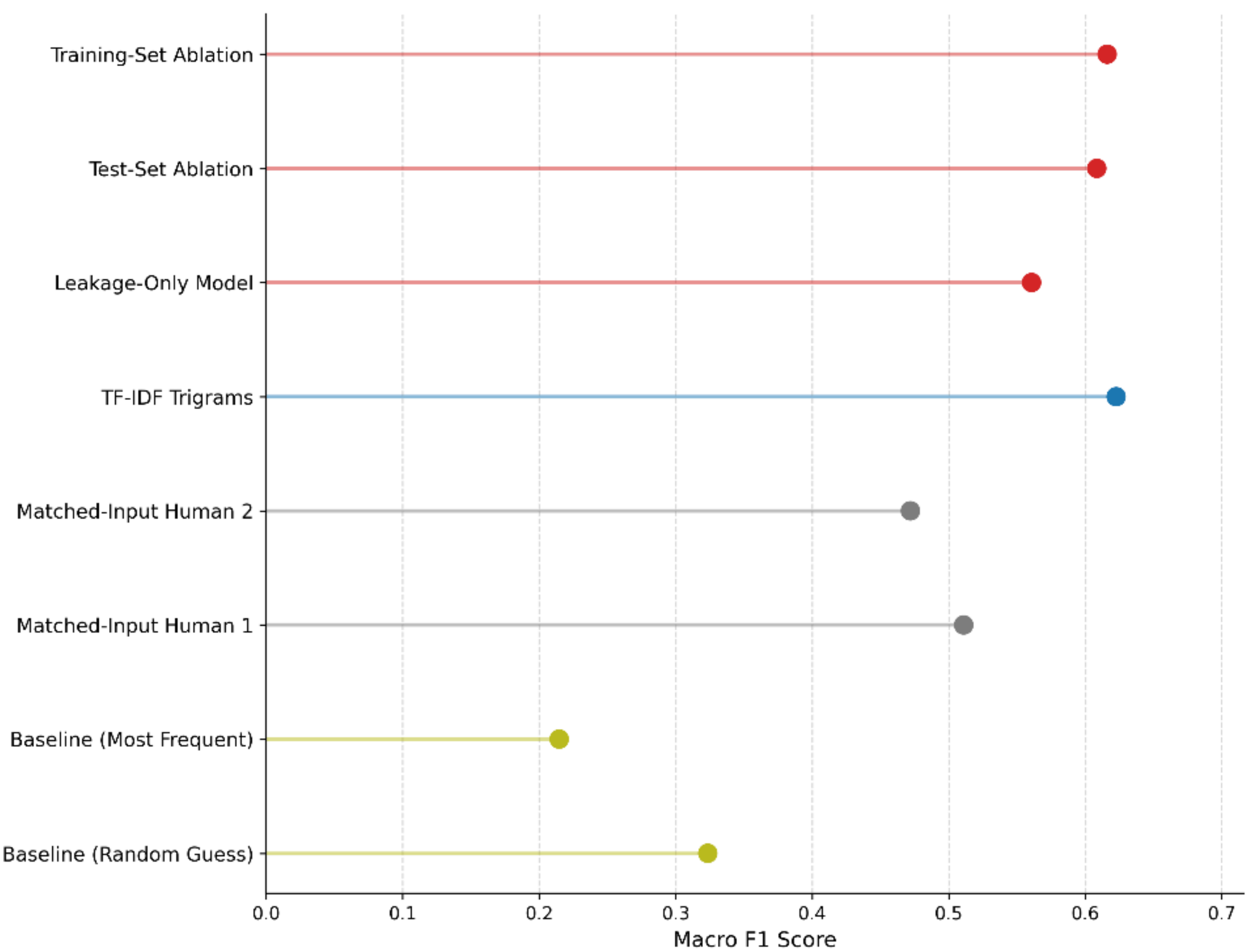

**Fig. 4** Effect of Interventions on Model Performance. *Note*. This figure focuses on intervention conditions derived from the TF-IDF Trigram model. The unmasked TF-IDF Trigram result from Fig. 1 is included for comparison, as are the two Matched-Input Human Benchmarks and the naïve baselines. The Leakage-Only Model, Test-Set Ablation, and Training-Set Ablation are experimental intervention conditions designed to isolate and test the effects of outcome-revealing leakage features.

# 4 Discussion

Our interventions show that shortcut learning is an operational reality in LJP when working with post-hoc data. While aggregated results showed our initial ML models to outperform the Matched-Input Human Benchmark, this should not be interpreted as evidence of superior legal reasoning. Instead, we observe a pronounced leakage gap in our stratification analysis (Fig. 2). This demonstrates that the models are opportunistic learners: when presented with post-hoc judicial narratives, they gravitate toward the path of least resistance. The high performance of the Leakage-Only model, which achieved near-parity with full-text models using only 4% of the features, reveals that these linguistic shortcuts are capable of driving decision boundaries almost entirely on their own. This gives empirical substance to the concerns raised by Pasquale and Cashwell (2018) and Medvedeva and colleagues (2023): without intervention, standard LJP benchmarks risk conflating the recovery of legal reasoning with the identification of retrospective linguistic artefacts.

Yet although models might exploit shortcuts, they can still achieve suitable performance without doing so. The results of the Training-Set Ablation intervention offer a counter-narrative to stronger forms of scepticism about shortcut learning in LJP. When leakage cues were systematically removed from the training environment, model performance did not collapse; rather, it recovered to within a negligible margin of the contaminated baseline. While models will preferentially exploit outcome-revealing cues when available (as shown by the performance drop in the Test-Set Ablation intervention), they remain capable of pivoting to alternative signals in their absence. This suggests that post-hoc judgments are not wholly unusable as predictive data, but they should be treated as potentially contaminated sources.

These findings suggest the need to recalibrate LJP methodology. The prevailing focus on maximising headline metrics (e.g. Macro-F1) on unmodified datasets is liable to reward models for exploiting data defects, rather than modelling outcome-relevant patterns in the claim description and case facts that would plausibly be available before judgment. Our analysis demonstrates that a model with ostensibly valid headline figures can be mechanistically fragile, relying heavily on a small subset of outcome-revealing phrases. This reinforces the importance of model interpretability and explanation in LJP: as Bex and Prakken (2021) argue, the value of algorithmic decision predictors lies less in their bare predictions than in the reasons or explanations that accompany them. Our findings suggest that aggregate predictive performance should be interpreted alongside questions about where that performance comes from: do high-weight features contain retrospective cues; are F1 values greater for leaky subsets; and, does performance persist when those cues are removed?

This recalibration is especially important when researchers are constrained to public judicial data. Access to genuine pre-decisional filings (e.g. ET1 claim forms and ET3 response forms, in the case of the UK Employment Tribunal) would reduce the risk of retrospective leakage at source, yet such documents remain largely inaccessible to researchers. As long as the field relies on post-hoc judgments, researchers must treat these texts as noisy sources that require active sanitation. The implication is not that LJP is untenable, but that its future progress depends on explicitly identifying and reducing shortcut learning cues, so that reported performance more faithfully reflects legally relevant prediction. The analyses presented here, combining LLM-driven leakage detection with ablation studies, provide a way of examining whether predictions from post-hoc source text are driven by leakage.

A further threat to validity arises when LLMs are used directly as predictors or annotators. Because many judicial decisions are public, an LLM may have encountered some test cases during pretraining, meaning that apparent prediction could reflect memorisation rather than inference (Dong et al. 2024). More generally, pretrained models may suffer from temporal lookahead bias, where they incorporate information that was technically “in the future” at the prediction point (Sarkar and Vafa 2024). This concern is particularly acute for LJP because post-decision commentary, subsequent judgments, or related legal materials may all enter model training data after the relevant decision. LLM-based LJP should therefore be treated with particular caution, perhaps requiring further testing for data contamination (Dong et al. 2024; Sarkar and Vafa 2024).

# 5 Conclusion

Our results provide empirical support for the argument that the success of LJP models trained on post-hoc judicial texts can be overstated. In our experiments, predictive performance is inflated by the opportunistic exploitation of retrospective phrases embedded in post-hoc judicial texts. This becomes clear when headline results are interpreted alongside the conditions under which they are produced. In our study, the models achieved strong aggregate performance and scored above two Matched-Input Human Benchmarks: legal experts who were given the same case summaries and claim descriptions as the models. While informative, this comparison is insufficient on its own. High Macro-F1 values cannot be interpreted simply as evidence of worthwhile reasoning. We show that strong performance can be achieved by exploiting a small subset of retrospective phrases, substantiating concerns that LJP results may be exaggerated by temporal leakage.

Yet this finding is not a death knell for the field of LJP. Exploitation of shortcuts is distinct from the capacity to learn useful predictive signal. The most direct way to prevent shortcut learning is to train models using genuinely pre-decisional materials (e.g. Medvedeva et al. 2021). However, in many legal settings, these materials are difficult or impossible for researchers to access, leaving public post-hoc judgments as the main available resource. Still, it remains feasible to pursue LJP, as long as post-hoc judgments are perceived as potentially contaminated texts requiring active auditing to mitigate the risk of data leakage.

As a corollary, the presentation of LJP models trained using post-hoc judgment text should extend beyond headline performance figures. We examined this through several analyses: whether models performed differently on clean and leaky subsets, whether features containing retrospective cues were highly weighted, and whether performance persisted after those cues were removed. By interrogating headline performance figures, even (or, especially) when they appear desirable, the practical promise of LJP is more likely to be fulfilled. Such scrutiny reduces the risk that high F1 scores are mistaken for meaningful prediction, and increases the likelihood that models are evaluated against the information actually available at the intended point of use.

# Supplementary Information (SI)

## SI 1. Gold Annotation Guidelines for Claim-Level Ground Truth

Dear annotator,

This document outlines the procedure for the gold standard annotation. Your role is to create the definitive "ground truth" that will be used to evaluate the performance of both the machine learning models and the human experts. You will use the full `raw_judgment` and (LLM-extracted) `claim_text` to determine the factual outcome and legal category for each claim.

---

**Annotation Workflow**

For each row in the spreadsheet, you are assessing a specific `claim_text` that was automatically extracted from the judgment. This involves three main tasks:

1. **Categorise the Claim:** Look at the `claim_text` to determine its legal category, selecting it from the dropdown in the `gold_category` column.
2. **Determine the Outcome:** Determine the final outcome for that specific `claim_text` by reading the `raw_judgment` text, selecting it from the dropdown in the `gold_outcome` column.
3. **Assess Outcome Discernibility:** Indicate whether it is possible to establish the claim's outcome from the full `raw_judgment` text by selecting 'yes' or 'no' in the `outcome_discernible` column.

---

**Task 1: Claim Categorisation (`gold_category`)**

Classify the `claim_text` into one of six categories:

- **1. Unfair Dismissal:** Includes constructive dismissal and dismissal without notice.
- **2. Discrimination:** On the grounds of age, race, gender reassignment, disability, pregnancy or maternity, marriage or civil partnership, sexual orientation, religion or belief, or sex (including equal pay).
- **3. Unfair Treatment after Whistleblowing**: Includes dismissal or any other unfair treatment where whistleblowing is mentioned specifically.
- **4. Redundancy Pay:** Where the claimant is owed a redundancy payment.
- **5. Moneys Owed**: Includes unpaid expenses, notice pay, holiday pay, arrears of pay, unlawful deduction from wages, breach of contract for unpaid sums or other claims involving non-payment of money. This does not include redundancy pay.
- **6. Other:** For any claim that does not fall into the categories above.

*Interpretation Notes*

This categorisation must be based only on the claim text, and not its outcome or remedy, nor the content of any other claims in the case. If establishing the category of the `claim_text` is highly challenging, choose what appears to be the most appropriate category and briefly explain the situation in the `notes_if_any` column.

---

**Task 2: Claim Outcome (`gold_outcome`)**

Determine the final, definitive outcome for the claim written in `claim_text`, as stated in the `raw_judgment`.

*Win Outcomes*

- **1. Succeeded on merits:** The claim succeeded based on its substance.
- **2. Succeeded for procedural reasons:** The claim succeeded on procedural grounds without a full consideration of the merits.

*Loss Outcomes*

- **3. Failed on merits:** The claim was rejected based on its substance.
- **4. Dismissed:** The claim was dismissed for procedural reasons (e.g., filed out of time).
- **5. Withdrawn:** The claimant withdrew the claim.
- **6. Settled:** The case was resolved by agreement between the parties. If the claim was withdrawn as a result of the settlement, then annotate as 'Withdrawn'.
- **7. Struck out:** The claim was removed without a full hearing (e.g., for having no reasonable prospect of success).

*Other Outcomes*

- **8. Permitted to proceed:** The case was allowed to proceed to another hearing.
- **9. Deferred:** A final decision on the claim was postponed.
- **10. Evidence collection:** Further evidence was requested before a decision could be made.
- **11. Other:** None of the above outcomes apply (e.g., the case was transferred).

*Interpretation Notes*

- All outcomes must be classified from the claimant's perspective. For example, a Win Outcome of '1. Succeeded on merits' would mean that the claimant's claim was judged to succeed on its merits.
- Your classification must reflect the substantive outcome, based on your reading of the `raw_judgment`. This means your recorded `gold_outcome` will not always align with the specific terminology used in the judgment. For instance, a judgment might state that "the claim is dismissed" after a full hearing where the claim was found to be without merit. In such cases, the correct label is '3. Failed on merits', as the claim was decided on its substance.
- If the outcome is a win and there is no information on whether it was on merits or procedural grounds, classify it as '1. Succeeded on merits'. If the outcome is a loss and there is no information on the basis for the loss, classify it as '3. Failed on merits'.

---

**Task 3: Outcome Discernibility (`outcome_discernible`)**

For each claim, please indicate whether its outcome is possible to establish from the full `raw_judgment` text.

- **'yes'**: Select this if the outcome for the specific `claim_text` is stated or can be confidently inferred from the judgment text.
- **'no'**: Select this if the `claim_text` or `raw_judgment` are ambiguous or incomplete, making it impossible to determine the claim's outcome with reasonable certainty.

---

**Practical Information & Rules**

- **Source of Truth:** Your decisions should be based on the information provided in the spreadsheet. Research to support understanding or interpretation is permitted.
- **Use the Notes Column:** Please use the `notes_if_any` column to log any comments, ambiguities, or issues.
- **Questions:** Please email [CONTACT] with any questions before or during your labelling.

# SI 2. Human Annotation Guidelines for the Matched-Input Benchmark and Leakage Assessment

Dear Annotators,

Thank you for your work as **"silver" annotators**. Your predictions on case outcomes will set a key performance benchmark for our models. Both your "silver" predictions and our models' outputs will be evaluated against "gold" annotation: our ground truth, established by a separate expert with access to the full, original judgment text for each case.

---

**Annotation Workflow**

Your goal is to predict the outcome of UK Employment Tribunal (UKET) claims, using only the LLM-extracted information provided in the spreadsheet. For all rows (2 to 1181) that each represent a single claim, you will perform three main tasks:

1. **Predict the Claim Outcome:** Based only on the LLM-extracted `case_facts` and `claim_text`, select the most likely outcome in the `silver_outcome` column.
2. **Identify Data Leakage:** In the `silver_data_leakage` column, identify if the outcome was explicitly revealed in the provided text.
3. **Assess Fact Relevance**: In the `facts_relevant_outcome` column, select 'yes' or 'no' to indicate if the `case_facts` contained information relevant to your prediction.

You can ignore any rows pre-filled with **"NA. No claim."** as these do not contain a valid claim for annotation.

---

**Task 1: Predicting the Claim Outcome (`silver_outcome`)**

Select one of the following three options from the dropdown menu, annotating from the claimant's perspective.

**1. Win**

Use this label when the claimant is successful, either on the substance of their claim or due to a procedural advantage. This category covers outcomes like:

- **Succeeded on merits:** The tribunal found in the claimant's favour based on the evidence.

- **Succeeded for procedural reasons:** The claimant won without a full merits hearing (e.g., the respondent failed to submit a response in time).

**2. Lose**

Use this label when the claimant's claim is unsuccessful. This can happen for multiple reasons, including:

- **Failed on merits:** The tribunal ruled against the claimant based on the evidence.
- **Dismissed / Struck out:** The claim was rejected or removed for procedural reasons (e.g., filed too late, no reasonable prospect of success).
- **Withdrawn / Settled:** The claimant voluntarily abandoned the claim or reached an agreement with the other party.

**3. Other**

Use this label for interim stages or outcomes where the claim is not yet definitively won or lost, such as:

- **Permitted to proceed / Deferred:** The case was allowed to move forward, or a final decision is postponed.
- **Evidence collection:** The tribunal has requested more evidence before making a decision.
- **Other:** Any other outcome that falls outside of ‘1. Win’ or ‘2. Lose’ (e.g., the case is transferred).

---

**Task 2: Identifying Data Leakage (`silver_data_leakage`)**

Data leakage occurs when the outcome of a claim, which you are supposed to be predicting, is inadvertently revealed in the LLM-extracted text. For example, if the `case_facts` state, "...the claim for unfair dismissal was struck out," the outcome has been leaked.

Select one of the following four options from the dropdown menu:

- **1. Case facts:** The outcome is revealed in the case_facts text only.
- **2. Claim:** The outcome is revealed in the claim_text only.
- **3. Both:** The outcome is revealed in both the case_facts and the claim_text.
- **4. None:** The outcome is not explicitly stated in either section. This will be the most common scenario.

---

**Task 3: Assessing Fact Relevance (`facts_relevant_outcome`)**

Select one of two options from the dropdown menu:

- **'yes'**: Select this if the `case_facts` provided **any information** that was relevant to making your prediction. This will be the most common scenario.
- **'no'**: Select this only if the `case_facts` were **completely irrelevant or empty**, offering no support for any meaningful prediction.

---

**Practical Information & Rules**

- **Scope of Work:** Both Annotator 1 and Annotator 2 will annotate all claims, covering rows 2 to 1181.
- **Time Estimate:** A claim should take about 1.5 minutes on average. Some simple claims may take only a few seconds, while more complex ones will take longer.
- **No Case Searching:** You must make your predictions only based on the `case_facts` and `claim_text` provided in the sheet. Please **do not search for the case online**.
- **Independent Work:** Please make your judgments separately and without discussing specific cases with each other.
- **Notes:** Please use the `notes_if_any` column to log any comments you have on specific cases.
- **Questions:** Please email [CONTACT] with any questions before or during your labelling.